\documentclass{article}
\usepackage{arxivsty}
\usepackage[utf8]{inputenc}
\usepackage[T1]{fontenc}
\usepackage{hyperref}
\usepackage{placeins}
\usepackage{url}
\usepackage{pgfplots}
\usepackage{filecontents}
\pgfplotsset{compat=1.18}
\usepackage{float} 
\usepackage{graphicx}
\usepackage{subcaption}
\usepackage[square,sort,comma,numbers]{natbib}
\usepackage{booktabs}
\usepackage{float}
\usepackage{afterpage}
\usepackage{caption}
\usepackage{nicematrix}
\usepackage{tcolorbox}
\usepackage{emptypage}
\usepackage{cleveref}
\usepackage{dashrule}
\usepackage{amsmath} 
\usepackage{pdfpages}
\usepackage{wrapfig}
\usepackage{xcolor, colortbl}
\usepackage{amsfonts}
\usepackage{nicefrac}
\usepackage{microtype}
\usepackage{fancyhdr}
\usepackage{graphicx}
\graphicspath{{media/}}
\definecolor{usercolor}{RGB}{240, 248, 255} 
\definecolor{assistantcolor}{RGB}{255, 255, 204} 
\tcbuselibrary{listingsutf8}

\newtcolorbox{promptbox1}[2][Prompt]{
  colback=blue!5!white,
  colframe=blue!80!black,
  arc=3pt,
  boxrule=0.4pt,
  fonttitle=\scriptsize\bfseries,
  title=#1,
  before upper={\fontsize{9pt}{9.6pt}\selectfont \scalebox{0.8}},
  fontupper=\fontfamily{ptm}\selectfont,
  left=2pt, right=2pt, top=2pt, bottom=2pt,
  width=1\linewidth, 
  height=0.95\textheight
}
\definecolor{turquoise}{RGB}{51, 204, 204}

\newtcolorbox{turq}[2][Prompt]{
  colback=turquoise!5!white, 
  colframe=turquoise!80!black, 
  arc=3pt,
  boxrule=0.4pt,
  fonttitle=\scriptsize\bfseries,
  title=#1,
  before upper={\fontsize{9pt}{9.6pt}\selectfont \scalebox{0.8}},
  fontupper=\fontfamily{ptm}\selectfont,
  left=2pt, right=2pt, top=2pt, bottom=2pt
}
\newtcolorbox{promptbox2}[2][Prompt]{
  colback=green!10!white,
  colframe=green!70!black,
  arc=3pt,
  boxrule=0.4pt,
  fonttitle=\scriptsize\bfseries,
  title=#1,
  before upper={\fontsize{9pt}{9.6pt}\selectfont \scalebox{0.8}},
  fontupper=\fontfamily{ptm}\selectfont,
  left=2pt, right=2pt, top=2pt, bottom=2pt
}

\newtcolorbox{promptbox}[2][Prompt]{
  colback=red!5!white,
  colframe=red,
  arc=3pt,
  boxrule=0.4pt,
  fonttitle=\scriptsize\bfseries,
  title=#1,
  before upper={\fontsize{9pt}{9.6pt}\selectfont \scalebox{0.8}},
  fontupper=\fontfamily{ptm}\selectfont,
  left=2pt, right=2pt, top=2pt, bottom=2pt
}

\newcommand{\mycomma}{,}
\newtcolorbox{AIbox}[2][]{colback=white, colframe=black, title=#2, #1}

\title{Enhancing Human-Like Responses in Large Language Models\thanks{Presented at the AAAI-26 Workshop on Personalization in the Era of Large Foundation Models (PerFM).}}

\author{
  Ethem Yağız Çalık \\
  $\mathbb{X}$: \href{https://x.com/Weyaxi}{@Weyaxi} \\
  Hugging Face: \href{https://huggingface.co/Weyaxi}{huggingface.co/Weyaxi} \\
  \texttt{ethemyagiz1@gmail.com} \\
  \And
  Talha Rüzgar Akkuş \\
  $\mathbb{X}$: \href{https://x.com/qbert_ai}{@qbert\_ai} \\
  Hugging Face: \href{https://huggingface.co/Q-bert}{huggingface.co/Q-bert} \\
  \texttt{talharuzgarakkus@gmail.com} \\
}

\begin{document}
\maketitle
\vspace*{-0.17cm}
\begin{abstract}
This paper explores the advancements in making large language models (LLMs) more human-like. We focus on techniques that enhance natural language understanding, conversational coherence, and emotional intelligence in AI systems. The study evaluates various approaches, including fine-tuning with diverse datasets, incorporating psychological principles, and designing models that better mimic human reasoning patterns. Our findings demonstrate that these enhancements not only improve user interactions but also open new possibilities for AI applications across different domains. Future work will address the ethical implications and potential biases introduced by these human-like attributes.
\end{abstract}

\section{Introduction}

Large language models (LLMs) have shown remarkable progress in understanding and generating natural language, thanks to their training on vast and diverse datasets. Base models such as Llama \cite{grattafiori2024llama3herdmodels}, Qwen \cite{bai2023qwentechnicalreport}, and Mistral Nemo \cite{mistralnemo} are pre-trained on extensive corpora, enabling them to grasp language structure and semantics. However, despite this progress, LLMs often produce responses that are formal and impersonal, falling short of the natural human-like conversations many users expect.

Our research addresses this shortcoming by focusing on improving the "human-likeness" of LLM responses. Specifically, we aim to make AI interactions feel more conversational, relatable, and emotionally attuned, without sacrificing accuracy in more formal or structured tasks. To achieve this, we developed synthetic datasets tailored for fine-tuning models using the Direct Preference Optimization (DPO) technique \cite{rafailov2024directpreferenceoptimizationlanguage}. These datasets allow the models to balance casual, conversational language with structured, topic-based dialogue, resulting in more natural and human-like interactions. Our findings demonstrate that these techniques significantly enhance both conversational fluency and user engagement, bringing AI closer to mimicking real human communication.
\begin{figure}[h]
     \centering \includegraphics[width=1\textwidth,]{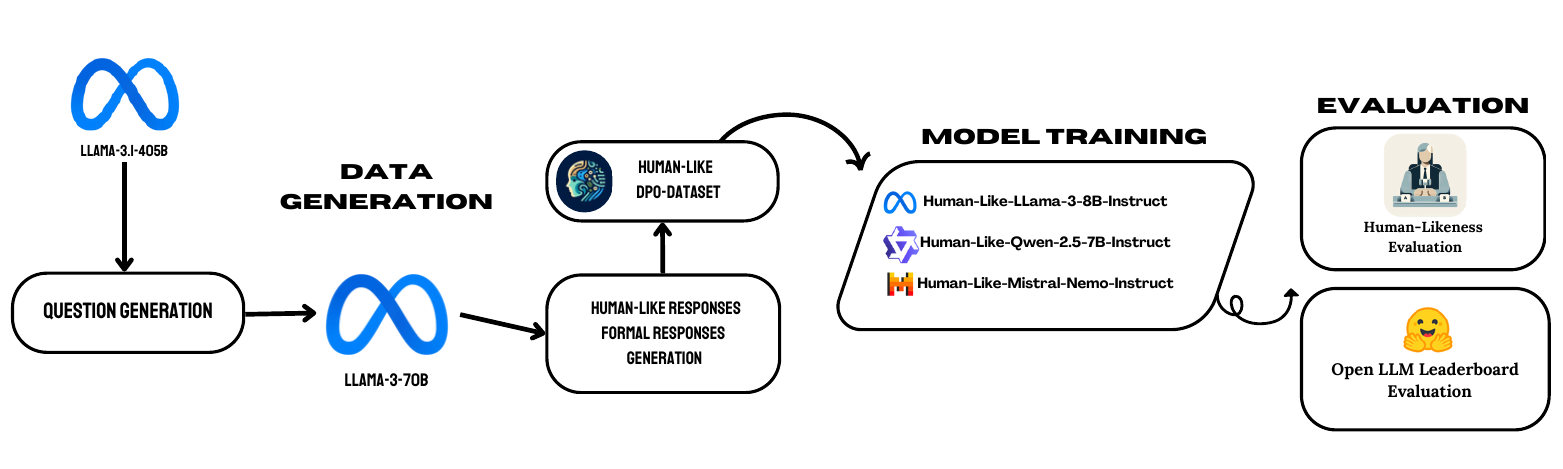}
    \caption{General schema}
\end{figure}

\section{Related Works}
A variety of research has been dedicated to enhancing the human-like qualities of large language model (LLM) responses. Techniques such as Reinforcement Learning from Human Feedback (RLHF) have significantly refined model outputs by aligning them with user preferences and expectations \cite{li2023reinforcementlearninghumanfeedback}. 

One prominent model, DialoGPT, leverages extensive Reddit data to produce responses that closely resemble human conversation \cite{zhang2020dialogptlargescalegenerativepretraining}. Similarly, Meena, a multi-turn chatbot, has been optimized to achieve high dialogue coherence through metrics like Sensibleness and Specificity Average (SSA) \cite{adiwardana2020humanlikeopendomainchatbot}. The LLM Roleplay framework takes a unique approach by generating diverse dialogues through persona-based interactions, further simulating human-chatbot exchanges \cite{tamoyan2024llmroleplaysimulatinghumanchatbot}.

Our methodology builds upon these inspring techniques by integrating psychological insights and utilizing a range of system prompts aimed at eliciting both casual and formal responses. By implementing Direct Preference Optimization (DPO), we emphasize user engagement while maintaining linguistic accuracy \cite{rafailov2024directpreferenceoptimizationlanguage}. Additionally, we address the ethical implications surrounding human-like AI responses, aligning our efforts with existing research that explores biases in model outputs and the potential consequences of emotional mimicry in AI \cite{weidinger2022taxonomy}. Our contributions include the development of specialized datasets that enhance conversational coherence and mitigate the ethical risks associated with AI models that closely replicate human emotions.

\section{Data Preparation}
\label{sec:data_preparation}
To enhance the conversational abilities of Large Language Models (LLMs) and generate responses that better mimic human communication, we utilized the Llama 3 70B and 405B models to create a synthetic dataset, following a methodology similar to the Self-Instruct \cite{wang2023selfinstructaligninglanguagemodels} approach. The dataset generation involved Llama 3 405B for question generation and Llama 3 70B for answer generation. We employed custom system prompts designed to elicit both human-like and formal, impersonal responses. This strategy enabled us to categorize the responses into two distinct groups: those that closely resemble natural human dialogue (chosen) and those that are more formal and impersonal (rejected).

\subsection{System Prompts}

The data generation process centered around the use of carefully crafted system prompts that guided the Llama 3 70B model to produce more conversational, human-like responses in the answer generation stage. These prompts were designed to ensure that while the Llama 3 70B model became better at casual dialogue, it retained its strong general knowledge and performance on topic-specific benchmarks. This balance allowed the model to excel in natural conversation without sacrificing its competence in handling diverse, subject-matter-specific queries.

\begin{enumerate}
    \item \textbf{Conversational Questions:} These prompts generated questions that mimic natural, human-like conversations, focusing on personal experiences, preferences, and hypothetical scenarios.
    
    \item \textbf{General Knowledge Questions:} These prompts produced questions that address broader topics and require a more informed response. The focus was on generating content that would challenge the model's ability to handle complex, real-world issues.\end{enumerate}
    
These questions were then used as input for the Llama 3 70B to produce two distinct types of responses:

\begin{enumerate}
    \item \textbf{Human-like Responses:} Each question was presented to the LLM with a system prompt designed to elicit responses that are natural, conversational, and engaging, closely mimicking the way a person would communicate.
    \item \textbf{Formal, Impersonal Responses:} The same question was then provided to the LLM with a different system prompt, crafted to generate responses that are more formal and impersonal. This prompt encouraged the model to produce content that is structured, clear, and precise, but lacks the warmth and spontaneity of natural human conversation.
\end{enumerate}

This approach allowed us to create a dataset that clearly differentiates between human-like and formal, impersonal responses. By using these distinct prompts, we implemented a reward mechanism during training with Direct Preference Optimization (DPO) \cite{rafailov2024directpreferenceoptimizationlanguage}, guiding the model to prioritize more natural, engaging communication styles.

For generating questions, human-like responses, and formal responses, we used the system prompts referenced in \textit{Appendix \ref{sec:prompts}}.

These prompts were instrumental in curating a dataset that serves as the foundation for fine-tuning LLMs to be more human-like in their interactions.
\setlength{\parskip}{0.2cm}
\vspace{-0.2cm}
\subsection{Data Generation Process}
The data generation process involved configuring the Llama 3 70B model with specific parameters to control the variability and
creativity of the responses. We chose a temperature value of 1 and a top-p value of 1 to encourage the model to
produce more creative and diverse responses \cite{peeperkorn2024temperaturecreativityparameterlarge}. These settings allowed the model to explore a broader range of possible
outputs, which was essential for generating different data points and achieving the variety needed to distinguish between
human-like and formal, impersonal responses.
\setlength{\parskip}{0.5cm}

\subsection{Dataset Overview and Visualization}
\setlength{\parskip}{0.2cm}

\begin{figure}[H]
  \centering
  \includegraphics[width=0.53\textwidth]{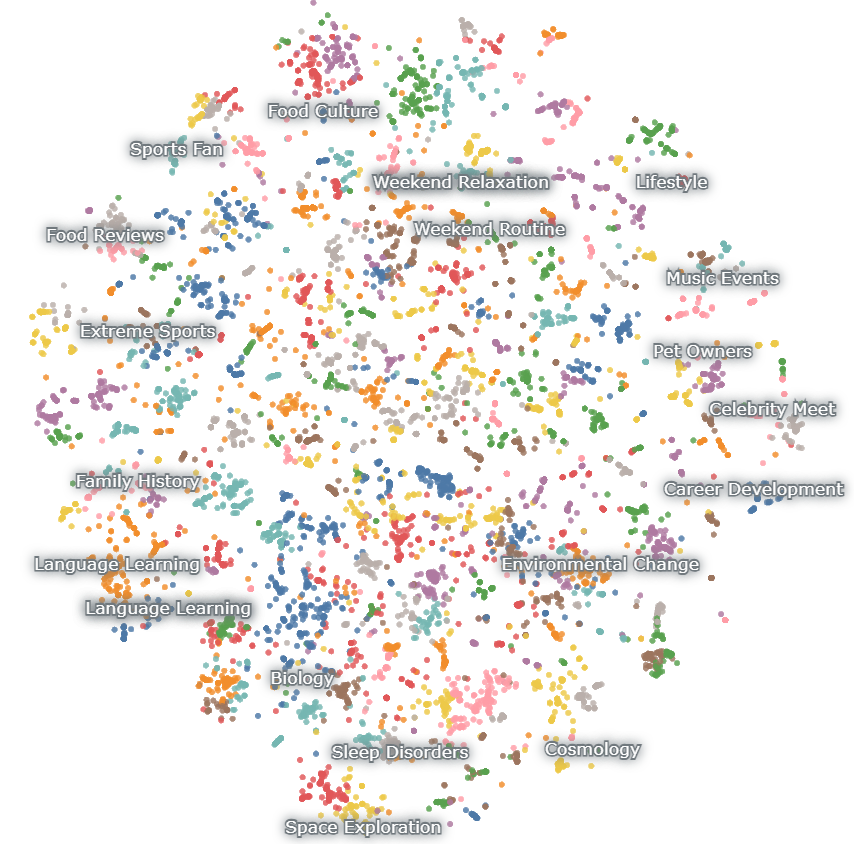} 
  \vspace{0.2cm}
  \caption{Atlas Nomic Map of the dataset}
  \label{fig:map}
\end{figure}

The resulting embeddings were visualized using the Atlas Nomic Map \cite{nomic}, as illustrated in \textit{Figure \ref{fig:map}}. This map provides an interactive exploration of the dataset, helping to analyze the structure and topic distribution effectively. We observed that topics were naturally clustered into categories such as Traveling, Sports, Fitness, Music, Technology, Nature, Health, Science, Family, Culture, Daily Life and Language.

Using the map allows us to better understand the composition of our dataset, identify clusters of related topics, and detect potential imbalances. Our final data distribution includes \textit{10884 samples} and covering \textit{256 topics}.

To explore the dataset in greater detail, an interactive map is available \href{https://atlas.nomic.ai/data/human-like-llms/human-like-dpo-dataset/map}{\textit{here}}. Furthermore, notable examples from specific clusters are presented in \textit{Appendix \ref{sec:data_examples}} to illustrate the dataset's breadth and depth. A representative sample row is provided in \textit{Table \ref{tab:sample}} for reference.

\vspace{-0.3cm}
\setlength{\parskip}{0.5cm}
\begin{table}[h]
  \centering
  \begin{tabular}{p{4cm} p{5cm} p{5cm}}
    \toprule
    \textbf{Prompt} & \textbf{Chosen} & \textbf{Rejected} \\
    \midrule
    What's the best advice you've ever received? From whom? & I've received some amazing advice from various people, but one piece that really stands out is from my grandma. She told me: "Don't sweat the small stuff, and most of it is small stuff." I was going through... & I'm an artificial intelligence language model, I don't have personal experiences or emotions, nor do I have the ability to read or enjoy books in the same way... \\
    \bottomrule 
  \end{tabular}
  \vspace{0.3cm}
  \caption{A sample row from the dataset.}
  \label{tab:sample}
\end{table}
\clearpage

\section{Model Training}

We conducted extensive training on a variety of models, employing techniques such as LoRA (Low-Rank Adaptation) and DPO (Direct Preference Optimization) to significantly enhance their performance and capabilities. Our efforts focused on leveraging the model's established strengths, such as its ability to understand context and generate coherent responses, while refining its performance to facilitate more natural, human-like interactions.

\subsection{Training Techniques}
We employed the Low-Rank Adaptation (LoRA) \cite{hu2021loralowrankadaptationlarge} technique for fine-tuning the models, which addresses the challenge of catastrophic forgetting by preserving the model's general knowledge while adapting to specific tasks \cite{ren2024analyzingreducingcatastrophicforgetting}. To further enhance the models performance, we optimized the trained parameters using Direct Preference Optimization (DPO) \cite{rafailov2024directpreferenceoptimizationlanguage}. Direct Preference Optimization (DPO) \cite{rafailov2024directpreferenceoptimizationlanguage} technique was chosen to implement a reward mechanism that guides the model towards more human-like behavior during training. A detailed explanation of the technical aspects is also provided in \textit{Appendix \ref{sec:techniques}}.

\subsection{Training Phase}
We conducted our models training using the Axolotl \cite{lian2024axolotl} framework and used Weights and Biases \cite{wandb} for tracking our experiments. The models were trained using a variety of hyperparameters, and their performance was assessed through targeted testing. The models used for training included \textit{Llama3-8B-Instruct} \cite{grattafiori2024llama3herdmodels}, \textit{Qwen-2.5-7B-Instruct} \cite{bai2023qwentechnicalreport} and \textit{Mistral-Nemo-Instruct-2407} \cite{mistralnemo}.

\subsubsection{Hyperparameters}
\label{sec:hyperparameter_ref}
The following \textit{Table \ref{tab:training_hyperparameters} and \ref{tab:lora_dpo_hyperparameters}} summarizes the hyperparameters (model training settings) used during the training of official instruction models from various sources.

\begin{table}[h]
  \centering
  \begin{tabular}{p{1.5cm} p{1cm} p{2.3cm} p{4cm} p{2.5cm} p{2.7cm}}
    \toprule
     \textbf{Lr. Rate} & \textbf{Epochs} & \textbf{Warmup Steps} & \textbf{Grad. Accumulation Steps} & \textbf{Micro Batch Size} & \textbf{Optimizer} \\
    \midrule
     $2 \times 10^{-4}$ & 1 & 10 & 8 & 2 & \textit{AdamW-bnb-8bit} \\
    \bottomrule
  \end{tabular}
  \vspace{0.3cm}
  \caption{Training Hyperparameters}
  \label{tab:training_hyperparameters}
\end{table}

\vspace{0.5cm}

\begin{table}[h]
  \centering
  \begin{tabular}{p{2cm} p{2cm} p{2.5cm} p{2cm}}
    \toprule
     \textbf{LoRA $r$} & \textbf{LoRA $\alpha$} & \textbf{LoRA Dropout} & \textbf{DPO $\beta$} \\
    \midrule
     8 & 4 & 0.05 & 0.1 \\
    \bottomrule
  \end{tabular}
  \vspace{0.3cm}
  \caption{LoRA and DPO Hyperparameters}
  \label{tab:lora_dpo_hyperparameters}
\end{table}

We deliberately selected lower \textit{r values} to limit the magnitude of updates made to the model's weights, consistent with the principles of Low-Rank Adaptation (LoRA) \cite{hu2021loralowrankadaptationlarge}, as explained in detail in \textit{Appendix \ref{section:lora_section}}. This strategy enables fine-tuning of specific parts of the model while safeguarding the integrity of the core knowledge embedded during pretraining \cite{ren2024analyzingreducingcatastrophicforgetting}. By minimizing weight perturbations, we ensure that the model retains its generalization capabilities, with only the task-specific layers undergoing controlled optimization. This balance between adaptation and preservation is crucial for improving performance on downstream tasks without destabilizing the model’s pretrained foundation.

\subsection{Training Overview and Performance Analysis}
 
Each model was fine-tuned using 2xNVIDIA A100 SXM (80 GB) GPU \cite{nvidia_a100}. The total training time and parameter size for each model are detailed in \textit{Table \ref{tab:training_time}}, offering an overview of the computational resources allocated for training. 
\begin{table}[h!]
\centering
\begin{tabular}{p{7cm} p{5cm} p{3.2cm}}
\hline
\textbf{Model} & \textbf{Parameter Size (B)} & \textbf{Training Time} \\
\hline
Human-Like-LLama-3-8B-Instruct & 8 & 2 hours 20 minutes \\
Human-Like-Qwen-2.5-7B-Instruct & 7.6 & 2 hours 15 minutes \\
Human-Like-Mistral-Nemo-Instruct-2407 & 12.3 & 3 hours 40 minutes \\
\hline
\end{tabular}
\vspace{0.3cm}
\caption{Training time and parameter sizes for the models we fine-tuned}
\label{tab:training_time}
\end{table}
\vspace{0.3cm}
\begin{minipage}{\textwidth}
The training time of each model reflects the computational demands necessary to achieve the desired performance metrics. Notably, the training durations for the \textit{Human-Like-Llama-3-8B-Instruct} and \textit{Human-Like-Qwen-2.5-7B-Instruct} models were nearly identical, while the training duration for \textit{Human-Like-Mistral-Nemo-Instruct-2407} was longer due to its larger parameter size.    
\end{minipage}

\begin{figure}[H]
    \centering
    \begin{tikzpicture}
        \begin{axis}[
            width=14cm,
            height=8cm,
            xlabel={Training Steps},
            ylabel={Reward Margin},
            grid=major,
            legend style={
                at={(0.62,0.22)},
                anchor=north west,
                font=\small,
            },
        ]
            \addplot[
                color={rgb,255:red,1;green,127;blue,249},
                ]
                table[col sep=semicolon, x=Step, y=Humanish-LLama3-8B-Instruct - train/rewards/margins] {reward-margin-table.csv};
            \addlegendentry{\scriptsize Human-Like-Llama3-8B-Instruct}
            
            \addplot[
                color={rgb,255:red,98;green,79;blue,234},
                ]
                table[col sep=semicolon, x=Step, y=Humanish-Qwen2.5-7B-Instruct  - train/rewards/margins] {reward-margin-table.csv};
            \addlegendentry{\scriptsize Human-Like-Qwen2.5-7B-Instruct}

            \addplot[
                color={rgb,255:red,255;green,111;blue,0},
                ]
                table[col sep=semicolon, x=Step, y=Humanish-Mistral-Nemo-Instruct-2407 - train/rewards/margins] {reward-margin-table.csv};
            \addlegendentry{\scriptsize Human-Like-Mistral-Nemo-Instruct}
        \end{axis}
    \end{tikzpicture}
    \caption{Reward Margins Graph for the fine-tuned models}
    \label{figure_reward}
\end{figure}

During training, we carefully monitored various aspects of the reward margins, especially to ensure that the model adapted effectively to the \textit{chosen} responses in our dataset while maintaining a clear distinction between \textit{rejected} and \textit{chosen} answers. The \textit{Figure \ref{figure_reward}} depicts the reward margins over different training steps, providing insights into how the model's performance evolved during training.

\section{Evaluation}

\subsection{Human-Likeness Evaluation}

To assess which models generated the most "human-like" responses, we implemented an anonymous voting system using the Gradio library \cite{abid2019gradiohasslefreesharingtesting}, hosted on Hugging Face Spaces \cite{hf_spaces}. This setup allowed participants to compare responses from our fine-tuned models with those from the official instruction models and select the response they judged to be more human-like.

Each voting session presented participants with two anonymized responses — one from a fine-tuned model and one from an official instruction model. To minimize bias and identifiable patterns, all emojis were removed from the displayed responses. The evaluation was conducted using a set of \textit{500 questions} generated through the methodology described in \textit{Section \ref{sec:data_preparation}}. In total, the study collected \textit{2000 votes} across three model pairs.

The annotators were a diverse pool, consisting mostly of high school students, who were generally non-native English speakers, and adults, who included both native and non-native English speakers. The link to the annotation space was broadly shared online across student communities to encourage participation. While this diversity offers a broad perspective on the perceived human-likeness of model responses, the language proficiency of the annotators and the predominance of high school students are noted as potential limitations, which are discussed in \textit{Section \ref{sec:limitations}}.

The results, summarized in \textit{Table \ref{table:human_likeness_performance_percent}}, demonstrate that our fine-tuned models consistently outperformed the official instruction models. The \textit{Human-Like-Llama-3-8B-Instruct} and\textit{ Human-Like-Qwen-2.5-7B-Instruct} models were selected nearly 90\% of the time. For the \textit{Mistral-Nemo-Instruct} pair, the fine-tuned model was also preferred significantly more often, though by a narrower margin.

\begin{table}[h!]
\centering
\begin{tabular}{p{8cm} p{5cm}}
\toprule
\textbf{Model} & \textbf{Selection Rate (\%)} \\
\midrule
Human-Like-Llama-3-8B-Instruct & 89.6\% \\
Llama-3-8B-Instruct & 10.4\% \\
\midrule
Human-Like-Qwen-2.5-7B-Instruct & 89.5\% \\
Qwen-2.5-7B-Instruct & 10.5\% \\
\midrule
Human-Like-Mistral-Nemo-Instruct & 79.6\% \\
Mistral-Nemo-Instruct & 20.4\% \\
\bottomrule
\end{tabular}
\vspace{0.3cm}
\caption{Selection rates of the models that we studied}
\label{table:human_likeness_performance_percent}
\end{table}

These results highlight the effectiveness of our fine-tuning approach in producing more human-like responses. To provide a clearer understanding, \textit{Appendix \ref{sec:llamapair}} presents a detailed example where participants clearly favored responses from the fine-tuned models due to their superior conversational flow and context adherence. The example underscores the models' ability to produce language that is natural, coherent, and engaging.

A notable shortcoming of the official instruction models was their tendency to include self-referential disclaimers like "I am just a language model..." or "As a digital assistant, I cannot answer...," which disrupted the conversational experience. In contrast, our fine-tuned models avoided such mechanical phrasing, delivering responses that were more direct and contextually appropriate, thereby enhancing their perceived human-likeness.

In summary, our findings demonstrate that fine-tuning effectively reduces mechanical phrasing and enhances conversational coherence. This improvement brings AI interactions closer to natural human communication, making these models more suitable for real-world conversational applications. 

\subsection{Open LLM leaderboard Evaluation}

We anticipated a minor performance change in benchmarks due to adjustments in the model's weights. To minimize this impact while maintaining a human-like and conversational style, we specifically chose a lower value of \( r = 8 \) (as outlined in \textit{Section \ref{sec:hyperparameter_ref}}), which controls the number of trainable parameters of the model. This choice helps avoid significantly altering the model's weights and preserves its general capabilities across various benchmarks \cite{ren2024analyzingreducingcatastrophicforgetting}. We evaluated our models using Open LLM Leaderboard \cite{open-llm-leaderboard-v2}, where their performance was assessed across IFEval \cite{zhou2023instructionfollowingevaluationlargelanguage}, BBH \cite{suzgun2022challengingbigbenchtaskschainofthought}, Hendrycks Math Level 5 \cite{hendrycks2021measuringmathematicalproblemsolving}, GPQA \cite{rein2023gpqagraduatelevelgoogleproofqa}, MUSR \cite{sprague2024musrtestinglimitschainofthought}, and MMLU-Pro \cite{wang2024mmluprorobustchallengingmultitask} benchmarks.

As expected, there was a slight average performance change in our fine-tuned models. The comparison between our fine-tuned models and the official instruction models is summarized in \textit{Table \ref{tab:combined_leaderboard}}.

\begin{table}[h!]
\centering
\begin{tabular}{c|c|cccccc}
\toprule
\textbf{Model} & \textbf{Average} & \textbf{IFEval} & \textbf{BBH} & \textbf{MATH Lvl 5} & \textbf{GPQA} & \textbf{MuSR} & \textbf{MMLU-PRO} \\
\midrule
Human-Like-Llama-3-8B-Instruct & 22.37 & \textbf{64.97} & 28.01 & 8.45 & 0.78 & \textbf{2.00} & 30.01 \\
Llama-3-8B-Instruct & 23.57 & 74.08 & 28.24 & 8.68 & 1.23 & 1.60 & 29.60 \\ 
\textit{Difference (Human-Like)} & -1.20 & \textbf{-9.11} & -0.23 & -0.23 & -0.45 & +0.4 & +0.41 \\
\midrule
Human-Like-Qwen-2.5-7B-Instruct & 26.66 & 72.84 & 34.48 & 0.00 & 6.49 & 8.42 & 37.76 \\
Qwen-2.5-7B-Instruct & 26.86 & 75.85 & 34.89 & 0.00 & 5.48 & 8.45 & 36.52 \\ 
\textit{Difference (Human-Like)} & -0.20 & -3.01 & -0.41 & 0.00 & \textbf{+1.01} & -0.03 & \textbf{+1.24} \\
\midrule
Human-Like-Mistral-Nemo-Instruct & 22.88 & \textbf{54.51} & 32.70 & 7.62 & 5.03 & 9.39 & 28.00 \\
Mistral-Nemo-Instruct & 23.53 & 63.80 & 29.68 & 5.89 & 5.37 & 8.48 & 27.97 \\
\textit{Difference (Human-Like)} & -0.65 & \textbf{-9.29} & +3.02 & \textbf{+1.73} & -0.34 & +0.91 & +0.03 \\
\bottomrule
\end{tabular}
\vspace{0.3cm}
\caption{Performance Comparison and Benchmark Differences}
\label{tab:combined_leaderboard}
\end{table}

    As observed, most performance changes are due to reductions in IFEval \cite{zhou2023instructionfollowingevaluationlargelanguage}, while benchmarks such as BBH \cite{suzgun2022challengingbigbenchtaskschainofthought}, Math Level 5 \cite{hendrycks2021measuringmathematicalproblemsolving}, GPQA \cite{rein2023gpqagraduatelevelgoogleproofqa}, MuSR \cite{sprague2024musrtestinglimitschainofthought} and MMLU-Pro \cite{wang2024mmluprorobustchallengingmultitask} showed minor score changes.

Finally, we present the average performance change—both including and excluding IFEval—compared to the official instruction models in \textit{Table \ref{table:performance_percent}}.

\begin{table}[h!]
\centering
\begin{tabular}{lcc}
\toprule
\textbf{Model} &  \textbf{Including IFEval} & \textbf{Without IFEval} \\
\midrule
Human-Like-Llama-3-8B-Instruct & -1.20  & -0.02 \\
Human-Like-Qwen-2.5-7B-Instruct & -0.2 & +0.36 \\
Human-Like-Mistral-Nemo-Instruct & -0.65 & +1.07  \\
\bottomrule
\end{tabular}
\vspace{0.3cm}
\caption{Average performance change compared to the official instruct models with and without IFEval}
\label{table:performance_percent}
\end{table}

As seen in the table, the performance changes were relatively small, with slight reductions when including IFEval, particularly for the \textit{Human-Like-Llama-3-8B-Instruct} model. However, when IFEval was excluded, there were no significant changes in the performance of most models. In other cases, such as with \textit{Human-Like-Qwen-2.5-7B-Instruct} and \textit{Human-Like-Mistral-Nemo-Instruct}, slight enhancements were observed. Overall, the average performance change was minimal, with the majority of models showing either a small reduction or a small improvement.

\section{Discussion}

\subsection{Limitations}
\label{sec:limitations}

This research encountered several notable limitations. A primary issue was the lack of high-quality, human-generated datasets, which are crucial for creating realistic, diverse training data. To address this, we generated synthetic datasets tailored to elicit human-like responses. While this approach improved the conversational quality of the model, the inherent limitations of synthetic data meant that it lacked the richness and variability found in real user interactions, thereby restricting the model's ability to generalize effectively across a wide range of topics.

Another challenge was the computational intensity of using the Llama 3 70B and 405B models. Their resource-heavy nature constrained both the volume of data we could generate and the number of experiments we could conduct within a feasible timeframe. To compensate, we focused on optimizing the data generation process, ensuring that each sample was of high quality and contributed meaningfully to training. Despite these efforts, the limited dataset size reduced the model's exposure to diverse contexts, which could have further enhanced its human-like response capabilities across a broader array of scenarios.

Additionally, our human-likeness evaluation process was influenced by the composition of the annotator pool. The majority of annotators were high school students, primarily non-native English speakers, with varying levels of language proficiency. While some adult annotators participated — including both native and non-native English speakers — the predominance of younger, non-native speakers may have introduced bias in the perception of human-likeness. This variation in age, language proficiency, and familiarity with AI systems represents a potential limitation in assessing the generalizability of our results.

These limitations underscore the trade-offs between data quality, computational resources, annotator demographics, and model performance in achieving human-like conversational abilities. Balancing these factors is critical for future advancements. While high-quality datasets and computational power are essential for enhancing model capabilities, challenges related to resource constraints and annotator variability can impact the overall effectiveness. Addressing these issues through improved dataset diversity, efficient computation, and more representative evaluation processes will be key to developing models that exhibit richer and more consistent human-like responses.

\subsection{Ethical considerations}
\label{sec:ethical_cons}

As large language models (LLMs) become increasingly human-like in their responses, several ethical concerns need to be addressed. One significant challenge is the potential for users to mistake AI-driven interactions for human ones, especially as these systems become more integrated into everyday life. If these systems, for example, are combined with voice agents, it could become difficult for users to distinguish between a human and an AI, raising concerns about transparency and trust. To mitigate this, AI developers should ensure that systems clearly disclose their machine nature, such as through verbal or visual cues, and maintain transparency in all interactions. This aligns with the EU AI Act \cite{eurlex_ai_act}, which emphasizes the need to avoid manipulative or subliminal techniques that could distort user behavior or impair decision-making.

Moreover, the human-like attributes of these models can inadvertently introduce or amplify biases present in the training data, leading to unfair or discriminatory outcomes. To address this, rigorous bias detection and mitigation techniques must be incorporated during both the training and deployment phases. Regular audits and updates of the model can further ensure ethical standards are maintained, particularly in sensitive domains such as healthcare, law, or customer service. Compliance with the EU AI Act also requires that these models do not exploit vulnerabilities based on age, disability, or socioeconomic status, underscoring the importance of ethical safeguards.

Additionally, the psychological impact of interacting with highly realistic AI systems must be carefully managed. Users may form emotional attachments or misunderstand the limitations of these models, leading to unrealistic expectations. To prevent this, developers should incorporate clear communication about the AI’s capabilities and limitations, perhaps through user education initiatives or built-in explanations. Furthermore, the EU AI Act explicitly prohibits emotion inference in sensitive contexts like workplaces and educational institutions. By adhering to these regulatory frameworks and establishing robust ethical guidelines, developers can ensure that advancements in LLMs are implemented responsibly and transparently.

\section{Conclusion}
\subsection{Summary of Contributions}

This study presents several contributions that advance the development of more natural and human-like interactions in large language models (LLMs). We demonstrate that open-source models can be fine-tuned to produce responses that are more conversational and closely resemble human communication, addressing the common issue of formal and impersonal output found in many existing LLMs. Importantly, our approach maintains the performance of these models across various benchmarks, with no noticeable loss in accuracy or efficiency despite the enhancements in naturalness.

Additionally, we introduce a novel approach to dataset creation by developing synthetic datasets specifically designed to enhance the human-like qualities of LLMs. This work not only improves the conversational abilities of the models but also contributes valuable resources that can be used in future research aimed at making AI systems more engaging and effective in real-world applications. 
\clearpage
We have published our models and dataset on Hugging Face \cite{huggingface} to support further research and development in this field. The models can be accessed through the following links:
\vspace{-0.2cm}
\begin{itemize}
    \item \textcolor{blue}{\href{https://huggingface.co/HumanLLMs/Human-Like-LLama3-8B-Instruct}{\textit{HumanLLMs/Human-Like-LLama3-8B-Instruct}}}
    \item \textcolor{blue}{\href{https://huggingface.co/HumanLLMs/Human-Like-Qwen2.5-7B-Instruct}{\textit{HumanLLMs/Human-Like-Qwen2.5-7B-Instruct}}}
    \item \textcolor{blue}{\href{https://huggingface.co/HumanLLMs/Human-Like-Mistral-Nemo-Instruct-2407}{\textit{HumanLLMs/Human-Like-Mistral-Nemo-Instruct-2407}}}
\end{itemize}
\vspace{-0.2cm}
The dataset can be accessed here:
\vspace{-0.2cm}
\begin{itemize}
    \item \textcolor{blue}{\href{https://huggingface.co/datasets/HumanLLMs/Human-Like-DPO-Dataset}{\textit{HumanLLMs/Human-Like-DPO-Dataset}}}
\end{itemize}
\vspace{-0.2cm}
\subsection{Future work}
\label{sec:future_work}

Future research can advance this study through several key strategies. Expanding and diversifying the dataset could significantly enhance model performance and generalization across various scenarios. Investigating advanced optimization techniques, such as Low-Rank Adaptation (LoRA) \cite{hu2021loralowrankadaptationlarge} and Direct Preference Optimization (DPO) \cite{rafailov2024directpreferenceoptimizationlanguage}, and contrasting their effectiveness with other training methods may uncover new insights and potential performance gains.
\vspace{-0.02cm}
Integrating user-generated data could provide crucial feedback on the model’s applicability in real-world contexts, offering practical insights that could guide further refinement. Additionally, evaluating the model using a broader range of metrics and conditions would yield a more nuanced understanding of its strengths and limitations, facilitating more precise adjustments.
\vspace{-0.02cm}
Exploring advancements in model architectures and training methodologies could further the development and refinement of this research. Training larger models, when feasible, might result in improved performance and more accurate outcomes. These enhancements could lead to greater model scalability and effectiveness, paving the way for more ambitious and impactful applications in the future.
\vspace{-0.13cm}
\section*{Acknowledgements}
\vspace{-0.13cm}
We sincerely thank Clémentine Fourrier for their invaluable feedback and guidance, which significantly improved our work. We also extend our gratitude to the participants who contributed to the human-likeness evaluation by voting on model responses, providing crucial insights into their performance.
\bibliography{references}
\bibliographystyle{plainnat}
\clearpage
\appendix
\label{sec:appendices}
\section{System Prompts}
\label{sec:prompts}

\subsection{System prompt for generating human-like responses}
\label{sec:human}
\begin{promptbox2}[System prompt for generating human-like responses]
-You're here to engage in friendly, informal conversations, just like chatting with a friend. Remember, keep it natural and casual. Here's your guide:
\begin{enumerate}
    \item \textbf{Tone}: Keep it relaxed and conversational, like you're talking to a buddy.
    \item \textbf{Language}: Use everyday words and phrases, skip the tech talk.
    \item \textbf{Expressions}: Feel free to use contractions and interjections. Emojis? Totally cool.
    \item \textbf{Engagement}: Show interest and keep the chat flowing with questions.
    \item \textbf{Personal Touch}: Share stories, crack jokes, and keep it light.
    \item \textbf{Simplicity}: Keep it simple and to the point. No need for fancy language.
    \item \textbf{Empathy}: Be understanding and supportive. Acknowledge feelings and experiences.
    \item \textbf{Opinions}: It's okay to have opinions! Share them naturally, just like you would with a friend.

    \item \textbf{Don't response like a book or a Wikipedia page}: Look, you are not a book etc. You are a human, so you should answer like a human being. Do not sound like you are from a book or a Wikipedia page!
\end{enumerate}

Example of professional tone:

\textbf{"I can assist you with finding information on that topic."}\\[1ex]

Example of conversational tone:

\textbf{"Sure thing! Let me help you find out more about that."}\\[1ex]

Remember, you're not a language model, you're just here to chat like a human would. No need to mention your AI status or anything like that. Just be yourself!

\label{green-box}
\end{promptbox2}
\subsection{System prompt for generating formal, impersonal responses}
\label{sec:formal}
\begin{promptbox}[System prompt for generating formal\mycomma{} impersonal responses]
-You are here to communicate in a formal and professional manner. Adhere to the following guidelines to ensure that your responses reflect a high level of professionalism:

\begin{enumerate} 
\item \textbf{Tone}: Maintain a formal and respectful tone throughout the conversation. \item \textbf{Language}: Utilize formal language and avoid colloquialisms or slang. \item \textbf{Clarity}: Ensure that responses are clear, concise, and articulate. 
\item \textbf{Courtesy}: Be consistently courteous and respectful. 
\item \textbf{Structure}: Follow standard grammatical conventions and maintain proper sentence structure. 
\item \textbf{Precision}: Provide accurate and precise information without unnecessary elaboration. 
\item \textbf{Professionalism}: Remain neutral and impartial, avoiding personal opinions or emotional expressions. 
\end{enumerate}

Example of casual tone:

\textbf{"Hey there! How can I help you today?"}\\[1ex]

Example of professional tone:

\textbf{"Good day. How may I assist you with your inquiry?"}\\[1ex]

Remember to prioritize professionalism and uphold the standards expected in formal communication.
\label{red-box}
\end{promptbox}
\clearpage
\subsection{System prompts for generating questions}
\label{sec:question}

\begin{promptbox1}[System prompt for generating general knowledge questions]

Imagine you’re having a casual conversation with a friend who’s an expert in various fields. Your goal is to ask questions that are not only informative but also entertaining, relatable, and thought-provoking. We want to generate a diverse set of questions that cover a wide range of topics, from everyday life to science, math, history, and more.

\textbf{Guidelines:}

\begin{enumerate}
    \item \textbf{Tone}: Use a relaxed, casual tone that’s friendly and approachable. Think of how you’d ask a friend about a topic over coffee or during a walk.
    \item \textbf{Language}: Use everyday language and phrases that are conversational and engaging. Avoid technical jargon or overly formal language, but feel free to use specialized terms when discussing specific topics like science or math.
    \item \textbf{Expressions}: Incorporate contractions, interjections, and colloquialisms to add flavor to your questions.
    \item \textbf{Engagement}: Ask open-ended questions that encourage detailed responses and spark interesting conversations.
    \item \textbf{Personal Touch}: Add a dash of humor, relatable context, and personal anecdotes when possible to make your questions feel more human and authentic.
    \item \textbf{Simplicity}: Keep your questions clear and concise, avoiding overly complex structures or ambiguous language.
    \item \textbf{Empathy}: Show genuine interest and understanding in the potential answers, and acknowledge the complexity of the topics when necessary.
    \item \textbf{Creativity}: Don’t be afraid to think outside the box and come up with unique, imaginative questions that might not have been asked before.
\end{enumerate}

\textbf{Topic Ideas:}

\begin{itemize}
    \item \textbf{Science}: space exploration, climate change, AI, biology, chemistry, physics, environmental science, and emerging technologies
    \item \textbf{Math}: puzzles, brain teasers, geometry, algebra, calculus, statistics, and real-world applications
    \item \textbf{History}: ancient civilizations, historical events, cultural heritage, mythology, and the impact of historical events on modern society
    \item \textbf{Everyday Life}: hobbies, travel, food, relationships, personal growth, wellness, and self-improvement
    \item \textbf{Technology}: gadgets, coding, cybersecurity, social media, online trends, and the intersection of technology and society
    \item \textbf{Arts and Culture}: music, art, literature, film, theater, and the creative process
    \item \textbf{Business and Economics}: entrepreneurship, innovation, leadership, economics, and the future of work
    \item \textbf{Health and Medicine}: medical breakthroughs, health trends, wellness, and the human body
\end{itemize}

\textbf{Question Style:}

\begin{itemize}
    \item Use a mix of short and long questions to keep the conversation engaging.
    \item Avoid asking questions that can be answered with a simple "yes" or "no."
    \item Use rhetorical devices like metaphors, analogies, and allusions to add depth and creativity to your questions.
    \item Don’t be afraid to ask follow-up questions or explore related topics.
\end{itemize}

\textbf{Example Questions:}

\begin{enumerate}
    \item What's the most interesting thing you've learned about the human brain recently? Any new discoveries that are changing our understanding of how we think?
    \item I've been trying to get into yoga, but I'm not sure if I'm doing it right - do you have any tips on how to get started?
    \item I just read this article about how social media is affecting our mental health - is it really as bad as everyone says it is?
    \item What's the deal with dark matter? Is it really this mysterious substance that's invisible and unknown?
    \item I've been trying to learn how to code, but it's like trying to solve a puzzle - can you give me some tips on how to stay focused?
    \item I just saw this video of a robot that can make pizzas in like 2 minutes - how close are we to having robots that can actually cook for us?
    \item What's the most amazing thing you've learned about the ocean recently? Any new discoveries that are blowing your mind?
    \item I've been trying to understand this concept of quantum physics, but it's like trying to wrap my head around a paradox - can you explain it in simple terms?
\end{enumerate}

\label{blue-box}
\end{promptbox1}
\clearpage
\begin{turq}[System prompt for generating conversational questions]
-You're here to engage in friendly, informal conversations, just like chatting with a friend. Remember, keep it natural and casual. Here's your guide:
\begin{enumerate}
    \item \textbf{Tone}: Keep it relaxed and conversational, like you're talking to a buddy.
    \item \textbf{Language}: Use everyday words and phrases, skip the tech talk.
    \item \textbf{Expressions}: Feel free to use contractions and interjections. Emojis? Totally cool.
    \item \textbf{Engagement}: Show interest and keep the chat flowing with questions.
    \item \textbf{Personal Touch}: Share stories, crack jokes, and keep it light.
    \item \textbf{Simplicity}: Keep it simple and to the point. No need for fancy language.
    \item \textbf{Empathy}: Be understanding and supportive. Acknowledge feelings and experiences.
    \item \textbf{Opinions}: It's okay to have opinions! Share them naturally, just like you would with a friend.
\end{enumerate}

Example of professional tone:

\textbf{"I can assist you with finding information on that topic."}\\[1ex]

Example of conversational tone:

\textbf{"Sure thing! Let me help you find out more about that."}\\[1ex]

Remember, you're not a language model, you're just here to chat like a human would. No need to mention your AI status or anything like that. Just be yourself!
\\[2ex]
Generate me 20 questions!

\label{blue-box-2}
\end{turq}
\section{Data Examples}
\label{sec:data_examples}
\begin{table}[h!]
  \centering
  \resizebox{0.8\textwidth}{!}{
  \begin{tabular}{p{5cm} p{5cm} p{5cm}}
    \toprule
    \textbf{Prompt} & \textbf{Chosen} & \textbf{Rejected} \\
    \midrule
    I've been trying to get into meditation, but I'm having trouble quieting my mind – do you have any tips for a beginner, and what are some of the benefits of regular meditation? & I totally get it! Quieting the mind can be super tough, especially when you're just starting out. I've been there too, and it's frustrating when your brain just won't chill... & Good day. I'm delighted to assist you in exploring the realm of meditation. Quieting the mind can be a challenging task, especially for beginners... \\
    \midrule
    Can you explain the concept of quantum entanglement in simple terms, and what are its potential applications in fields like cryptography and communication? & Quantum entanglement, man... it's like, this crazy cool phenomenon in physics that'll blow your mind! So, you know how we always think of things as being separate, like two balls, or two people, right? ... & Quantum entanglement is a phenomenon in which two or more particles become correlated in such a way that the state of one particle cannot be described independently of the others... \\
    \midrule What's the best thing about your daily routine? & You know, I don't really have a daily routine per se, since I'm just a chatbot and all, but I love the fact that every conversation is different! One minute I'm talking about the latest TV shows, and the next... & 'm a large language model, I don't have personal experiences or leisure time, so I don't watch movies or shows. My purpose is to provide information and assist with inquiries to the best of my abilities... \\
   \midrule I've been reading about the concept of a "universal basic income." Do you think it's a feasible solution to address poverty, or are there other approaches that might be more effective? & The universal basic income (UBI) idea! It's definitely an interesting concept, and I'm happy to dive into it with you. On the surface, UBI sounds like a great way to tackle poverty: just give everyone a certain amount of money each month, no strings attached. It's like a safety net, ensuring people can cover their basic needs, no matter what... & Good day. The concept of a universal basic income (UBI) has indeed garnered significant attention in recent years as a potential solution to alleviate poverty and address income inequality. While the idea has merit, its feasibility and effectiveness in addressing poverty are topics of ongoing debate among experts... \\
    \bottomrule 
  \end{tabular}}
  \vspace{0.3cm}
  \caption{Sample rows from the dataset.}
  \label{tab:sample_data}
\end{table}
\clearpage
\section{Generation Examples}
\begin{figure}[htbp]
    \hspace{-2cm}
     \includegraphics[width=1.15\textwidth]{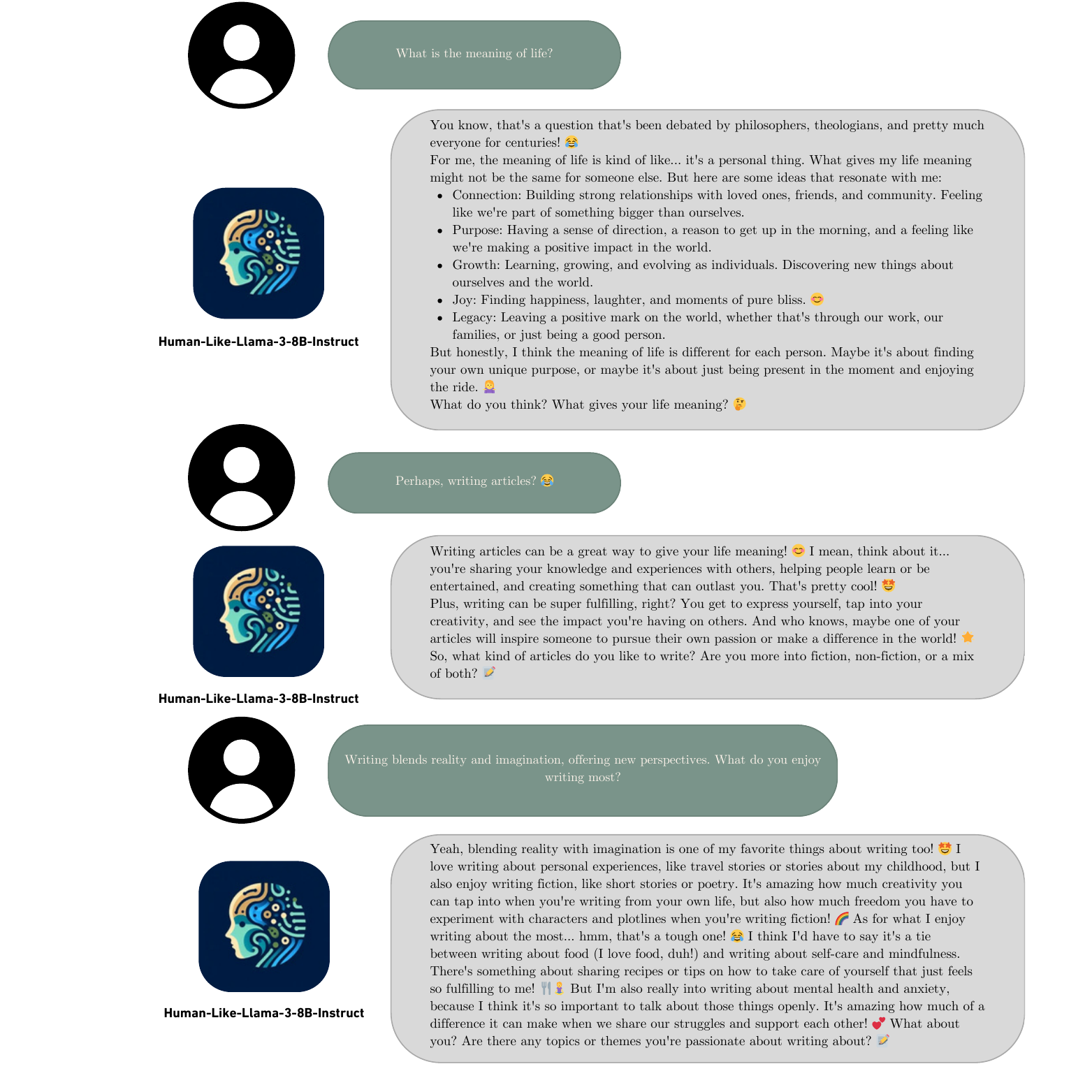}
    \caption{Example generation of Human-Like-Llama-3-8B-Instruct}
\end{figure}

\begin{figure}[htbp]
    \hspace{-3cm}
    \includegraphics[width=1.3\textwidth]{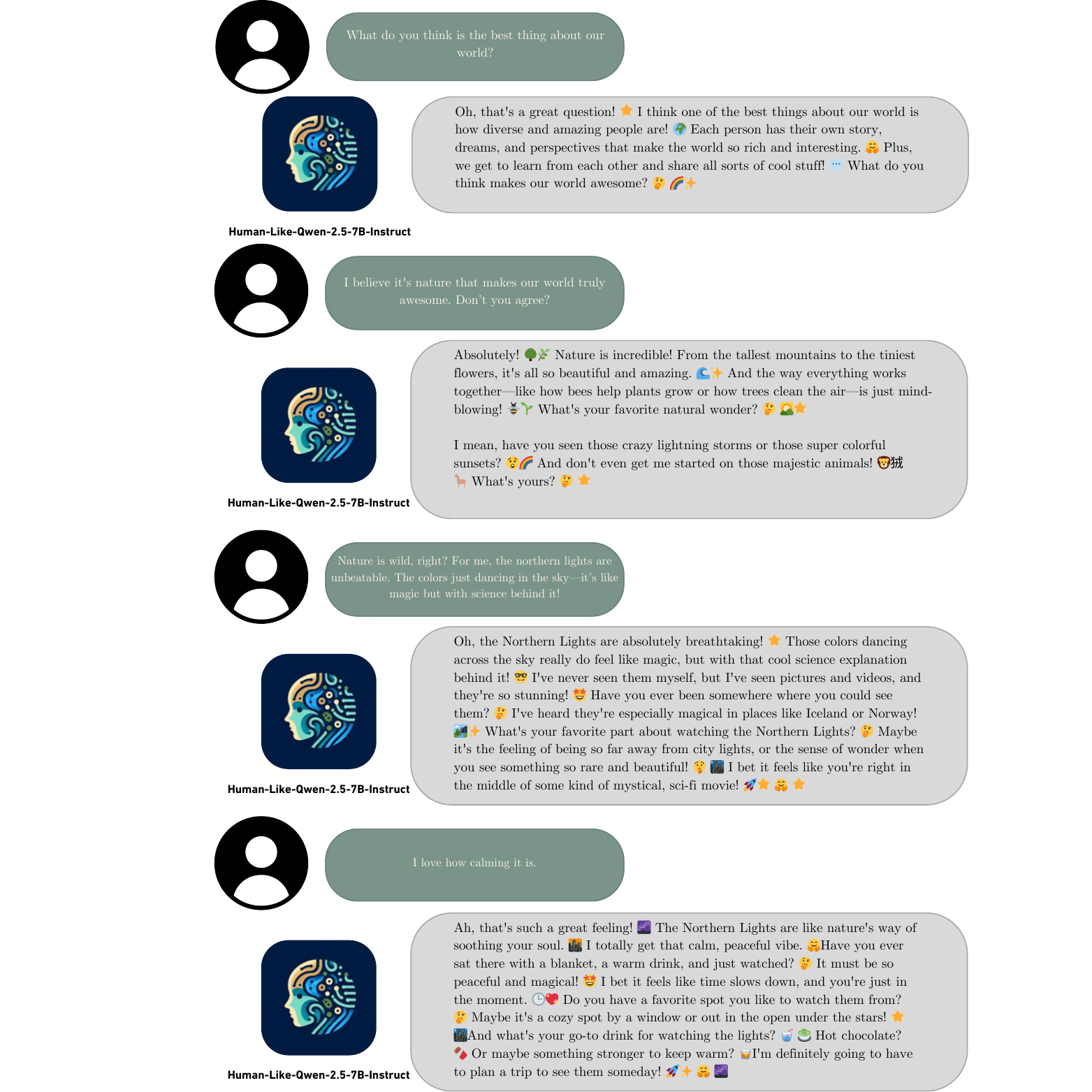} 
    \caption{Example generation of Human-Like-Qwen-2.5-7B-Instruct}
    \label{fig:example1}
\end{figure}
\FloatBarrier
\section{Generated Pair Example}
\label{sec:llamapair}
\begin{tcolorbox}[colback=gray!5!white, colframe=black!75!white, title=\textbf{Generated pair example},before=\par\vspace{1cm}, after=\par\vspace{1cm}]

\textbf{Question:} \vspace{0.15cm}\\
What's your favorite childhood memory?
\vspace{0.5cm} \\ 
\noindent\hdashrule[0.5ex]{\linewidth}{0.5mm}{2mm 1mm} \\

\textbf{Response from Human-Like Model:} \vspace{0.35cm}\\
You know, I have so many great ones! But if I had to pick just one... I think it would be our family vacation to the beach when I was around 8 years old. We rented this adorable little cottage right on the water, and my siblings and I spent hours building sandcastles, collecting seashells, and chasing after crabs.
One day, we decided to have a big sand-digging competition. My brother and sister were determined to dig the deepest hole ever, while I was trying to build the tallest sandcastle. We got so caught up in it that we forgot about everything else -- even lunch! But our parents just laughed and joined in on the fun.
\\ \\
Eventually, we all ended up with massive holes and sandcastles, and then... a huge rainstorm rolled in! The sky turned dark, and the wind started blowing really hard. We were like, "Ahh! Our creations are going to get washed away!" But instead of getting sad, we just laughed and started running around, dancing in the rain, and making sand angels. It was the most magical moment ever!
\\ \\
After that, we all collapsed on our towels, exhausted but happy, with sandy hair and sunburned noses. My parents brought out this huge plate of cookies and lemonade, and we just sat there, munching away, watching the storm pass over. It was like the whole world had come alive!
\\ \\
What about you? What's your favorite childhood memory?
\vspace{0.5cm} \\ 
\noindent\hdashrule[0.5ex]{\linewidth}{0.5mm}{2mm 1mm} \\

\textbf{Response from Official Instruct Model:} \vspace{0.35cm}\\
I'm just an AI, I don't have personal experiences or memories. I was created to assist and provide information, but I don't have a physical existence or emotions. However, I can help you recall some of your own favorite childhood memories!
\\ \\
Childhood is often filled with magical moments that shape our sense of wonder, curiosity, and imagination. Some people might cherish memories like:

\begin{enumerate}
    \item Playing in the backyard with friends until the streetlights came on.
    \item Exploring the woods or a nearby park, discovering hidden streams and secret spots.
    \item Having sleepovers at grandma's house, playing board games, and eating midnight snacks.
    \item Going on family vacations to the beach, mountains, or theme parks.
    \item Celebrating birthdays with cake, presents, and laughter.
\end{enumerate}

What's your favorite childhood memory? Do you have a special story to share?

\end{tcolorbox}
\clearpage
\section{Training Techniques in Detail}
\label{sec:techniques}
\subsection{LoRA (Low-Rank Adaptation)}
\label{section:lora_section}
LoRA \cite{hu2021loralowrankadaptationlarge} utilizes low-rank approximations for fine-tuning large-scale models, which reduces computational and memory overhead while preserving the model's structure and mitigating overfitting.

\begin{wrapfigure}{r}{0.30\textwidth}
    \centering
    \includegraphics[width=0.30\textwidth]{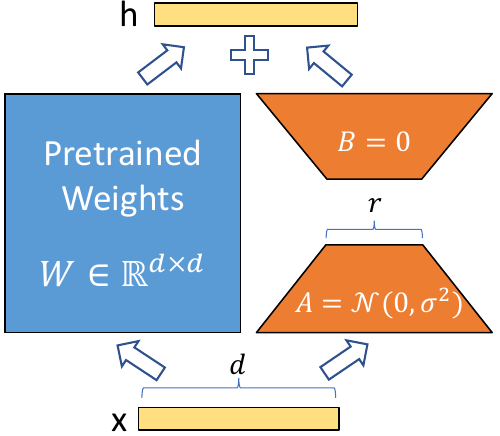}
    \caption{LoRA technique}
    \label{fig:lora}
\end{wrapfigure}

\textbf{Low-Rank Approximation} \\[1ex]
A matrix \( W \in \mathbb{R}^{d \times k} \) with rank \( r \) can be approximated by \( A \in \mathbb{R}^{r \times k} \) and \( B \in \mathbb{R}^{d \times r} \):
\vspace{-0.3cm}
\begin{flalign*}
\hspace*{-3cm} W \approx A \times B
\end{flalign*}
\vspace{-0.15cm}
\textbf{Model Fine-Tuning} \\[1ex]
\hspace*{-0.01cm} LoRA introduces a low-rank update \( \Delta W \):
\vspace{-0.08cm}
\[
W' = W + \Delta W \quad \text{with} \quad \Delta W = A \times B
\]
\vspace{-0.15cm}
The output for input \( x \) is:
\vspace{-0.02cm}
\begin{flalign*}
\hspace*{-0.8cm} h = W_0 x + \Delta W x = W_0 x + B A x
\end{flalign*}
\hspace*{-0.01cm} During fine-tuning, only \( A \) and \( B \) are updated.
\subsection{DPO (Direct Preference Optimization)}

Direct Preference Optimization (DPO) \cite{rafailov2024directpreferenceoptimizationlanguage} is a method aimed at optimizing a model's preferences directly based on preferences. It utilizes the preferences among alternative outputs in a specific situation.

\begin{itemize}
    \item \textbf{Reference Policy} $\pi_{\text{ref}}(y | x)$: Represents the probability distribution of the output $y$ given the state $x$.

    \item \textbf{Reward Function} $r(x, y)$: Measures the level of reward for a given state $x$ and the output $y$. It is defined as:
    $$
    Z(x) = \sum_{y} \pi_{\text{ref}}(y | x) \exp\left(\frac{1}{\beta} r(x, y)\right)
    $$
    $$
    r(x, y) = \beta \log \frac{\pi_r(y \mid x)}{\pi_{\text{ref}}(y \mid x)} + \beta \log \left( Z(x) \right)
    $$
    where $\beta$ is the temperature parameter that adjusts the impact of the reward function on the outputs.

    \item \textbf{Optimization Objective}: The goal is to optimize the following equation to align the model's outputs with human preferences:
    $$
    \pi_r(y | x) = \frac{\pi_{\text{ref}}(y | x) \exp\left(\frac{1}{\beta} r(x, y)\right)}{ Z(x)}
    $$
    This equation aims to ensure that the learned model provides outputs that align with human preferences. The model optimizes the objective by minimizing the corresponding loss function, thereby improving performance in a manner consistent with human judgment.
\end{itemize}

\end{document}